# How to fine-tune deep neural networks in few-shot learning?

Peng Peng*, Jiugen Wang

Faculty of Mechanical Engineering, Zhejiang University, Hangzhou, 310027, China

*Corresponding author: Peng Peng

E-mail: pengpzju@163.com

**Abstract:**

Deep learning has been widely used in data-intensive applications. However, training a deep neural network often requires a large dataset. When there is not enough data available for training, the performance of deep learning models is even worse than that of shallow networks. It has been proved that few-shot learning can generalize to new tasks with few training samples. Fine-tuning of a deep model is simple and effective few-shot learning method. However, how to fine-tune deep learning models (fine-tune convolution layer or BN layer?) still lack deep investigation. Hence, we study how to fine-tune deep models through experimental comparison in this paper. Furthermore, the weight of the models is analyzed to verify the feasibility of the fine-tuning method.

## 1 Introduction

In order to explore the potential of deep learning, Krizhevsky et al. proposed a deep convolution neural network, AlexNet in 2012 [1]. AlexNet won the first prize in the ImageNet competition and the top-5 test error rate is much lower than the second-best entry. After that, deep convolution neural networks have been widespread concern by researchers, and many more deep models have been proposed. Currently, deep convolution neural networks have become the most prevailing method in computer vision. Simonyan proposed the VGG network, which not only deepens the network depth but also greatly improves the nonlinear expression ability of the network [2]. Szegedy designed the inception module in convolution network and propose the inception network models [3-6]. He et al. proposed the residual network. The residual network greatly deepens the network layers of the model by using the skip connection [7]. Huang et al. proposed the DenseNet model, which establishes the dense connection between the front and back network layers, and also greatly deepens the number of network layers of the model [8]. From the above convolution network design development process, the network structure is getting deeper and deeper. The success of these deep models can be attributed to the improvement of computing power and the tremendous labelled data. However, many computer vision tasks lack sufficient labelled data. For example, the ferrograph images used for the wear detection of mechanical equipment in the industry are often insufficient. In this case, the performance of deep neural networks will be greatly degradative. Recently, few-shot learning has been proposed to tackle this problem [9]. Few-shot learning method can generalize to new tasks with few training samples. Fine-tuning a deep model is a simple and effective few-shot learning method. Yosinski et al. make a detailed study on the transferable portability of AlexNet [10]. However, there are few layers in the AlexNet and the BN (batch normalized) layer [4] used in mainstream networks is not included in AlexNet. BN layers enable each layer of the deep network to learn a similar data distribution, which can accelerate model convergence and prevent over-fitting of the model. However, recent studies show that BN layers may be related to the data domain and may not be generalizable to other domain [11]. Thus, how to deal with BN layers in fine-tuning when the target domain is different from the source domain?

Ferrography is a feasible wear detection technology in the industry. It can directly reflect the fault information of mechanical equipment in the form of ferrography image [12, 13]. However, the acquisition of ferrography image needs much time and manpower. Therefore, the amount of ferrography image is generally small. One or two ferrograph images can be collected when the fault is rare. Recently, fine-tuning methods have been applied to classify ferrograph images [14, 15]. However, these researches lack deep investigation of fine-tuning. Hence, we will carry out a large number of experiments on fine-tuning to explore whether fine-tuning can be used for ferrograph image recognition, and how to design a feasible and effective fine-tuning approach for ferrograph image classification. Specifically, the following problems will be studied in this research:

(1) Are fine-tuning effective? The data distribution of ferrograph image is quite different from that of the ImageNet. Will negative transfer [16] occur?

(2) Which fine-tuning approach is effective? Fine-tune all weights or partial parameters?

(3) Fine-tune what kind of layer? Fine-tune what kind of layer will achieve the best result?

(4) Are BN layers needed to be fine-tuned? Which BN layers need to be fine-tuned?

(5) Why fine-tuning is effective?

## 2 Method

We collected oil samples form mechanical equipment and made into ferrograph images. There are seven types of ferrograph images, namely background image, fatigue wear particle image, oxidized wear particle image, spherical wear particle image, fatigue wear particle and oxide wear particle image, fatigue wear particle and spherical wear particle image, and oxidation wear particle and spherical wear particle image in this study. We randomly selected 909 images as the training set and 881 images as the test set. Besides, 20 and 40 samples for each class are randomly sampled from the training set to obtain two other small data sets. The subsequent experiments in this research will be carried out on these three datasets. The training data set with 909 images is called FI909, FI140 and FI280 are extrapolated in this way.

Similar to [11] [17], we designed a metric based on KL divergence to analyze the difference of the weights of different fine-tuning models. We assume the weights in the deep models follow a Gaussian distribution with mean $\mu$ and variance $\sigma^2$. Then the KL divergence of the weights of two different fine-tuning models A and B would be:

$$KL(N(\mu_A, \sigma_A^2) \| N(\mu_B, \sigma_B^2)) = \int_x \frac{1}{\sqrt{2\pi}\sigma_A} e^{-\frac{(x-\mu_A)^2}{2\sigma_A^2}} \log \frac{\frac{1}{\sqrt{2\pi}\sigma_A} e^{-\frac{(x-\mu_A)^2}{2\sigma_A^2}}}{\frac{1}{\sqrt{2\pi}\sigma_B} e^{-\frac{(x-\mu_B)^2}{2\sigma_B^2}}} dx$$

$$= \log \frac{\sigma_B}{\sigma_A} + \frac{\sigma_A^2 + (\mu_A - \mu_B)^2}{2\sigma_B^2}$$

(1)

# 3 Experiment

We adopt the classical deep models DenseNet121, DenseNet201, ResNet18, ResNet50, and ResNet152 in this study.

## 3.1 Transfer weights and train FC layers

First, we will investigate whether transfer weights from pre-trained deep model and only train the FC (Full connected) layers can improve the accuracy of ferrograph image classification. Therefore, we carry out two groups of comparative experiments, namely, training the whole deep models from scratch and transfer weights from pre-trained deep models and only train the FC layers. When all models are trained from scratch, we find that the larger learning rate cannot make the models converge. Especially, gradient explosion occurs in the very deep models. Therefore, we set the learning rate to 0.001 and adopt Adam optimizer. When the weights are transferred and the FC layers are trained, we find the learning rate 0.001 also make the best results. Thus, the learning rate is also set to 0.001 in this case. The experimental results are shown in Table 1-3.

Table 1 Train from scratch VS transfer weights and train FC layers in FI140.

| Models | Train from scratch | Transfer weights and train FC layers |
| --- | --- | --- |
| DenseNet121 | 56.6402 | 58.6833 |
| DenseNet201 | 57.8888 | 56.0726 |
| ResNet18 | 51.0783 | 54.9376 |
| ResNet50 | 51.0783 | 59.0238 |
| ResNet152 | 50.5108 | 61.521 |

Table 2 Train from scratch VS transfer weights and train FC layers in FI280.

| Models | Train from scratch | Transfer weights and train FC layers |
| --- | --- | --- |
| DenseNet121 | 70.1476 | 62.4291 |
| DenseNet201 | 70.1476 | 63.4506 |
| ResNet18 | 59.2509 | 59.0238 |
| ResNet50 | 60.4994 | 63.4506 |
| ResNet152 | 56.6402 | 62.3156 |

Table 3 Train from scratch VS transfer weights and train FC layers in FI909.

| Models | Train from scratch | Transfer weights and train FC layers |
| --- | --- | --- |
| DenseNet121 | 86.7196 | 74.5743 |
| DenseNet201 | 86.4926 | 77.185 |
| ResNet18 | 81.3848 | 72.7582 |
| ResNet50 | 81.8388 | 74.6879 |
| ResNet152 | 81.7253 | 76.277 |

As shown in Table 1-3, when the quantity of the ferrograph image is small, transfer weights and only train the FC layers is effective. However, when the number of the ferrograph image is larger, training the deep models from scratch obtains better results, which means negative transfer occurs [16]. Moreover, we find that training the deep models from scratch can achieve more than 80% accuracy in FI909, which shows the classify task is relatively simple.

*3.2 Fine-tuning CNN layers or BN layers*

We find transferring weights and training the FC layers may not achieve good results in Section 3.1. In order to further explore the potential of fine-tuning, we will adopt two other fine-tuning approaches, that is fine-tune CNN layers and fine-tune BN layers, in this section. The learning rate in fine-tune CNN layers is set to 0.0001, while the learning rate in fine-tune BN layers is set into 0.01. We found that such a learning rate setting can make the network obtain better results. The experimental results are shown in Table 4-6.

Table 4. Fine-tune CNN layers or BN layers in FI140.

| Models | FC | CNN & FC | BN & FC | CNN, BN & FC |
|---|---|---|---|---|
| DenseNet121 | 58.6833 | 66.0613 | 66.9694 | 66.4018 |
| DenseNet201 | 56.0726 | 68.1044 | 69.2395 | 67.7639 |
| ResNet18 | 54.9376 | 64.8127 | 66.2883 | 64.5857 |
| ResNet50 | 59.0238 | 63.6776 | 66.8558 | 65.0397 |
| ResNet152 | 61.521 | 66.2883 | 65.1532 | 66.7423 |

Table 5. Fine-tune CNN layers or BN layers in FI280.

| Models | FC | CNN & FC | BN & FC | CNN, BN & FC |
|---|---|---|---|---|
| DenseNet121 | 62.4291 | 71.5096 | 83.0874 | 68.5585 |
| DenseNet201 | 63.4506 | 71.1691 | 82.7469 | 71.1691 |
| ResNet18 | 59.0238 | 71.5096 | 80.0227 | 70.2611 |
| ResNet50 | 63.4506 | 70.8286 | 81.0443 | 74.3473 |
| ResNet152 | 62.3156 | 72.9852 | 79.2281 | 72.1907 |

Table 6. Fine-tune CNN layers or BN layers in FI909.

| Models | FC | CNN & FC | BN & FC | CNN, BN & FC |
|---|---|---|---|---|
| DenseNet121 | 74.5743 | 91.2599 | 93.7571 | 90.2384 |
| DenseNet201 | 77.185 | 91.0329 | 93.4166 | 91.941 |
| ResNet18 | 72.7582 | 85.1305 | 90.3519 | 87.9682 |
| ResNet50 | 74.6879 | 88.1952 | 93.076 | 89.7843 |
| ResNet152 | 76.277 | 89.4438 | 92.2815 | 91.0329 |

As shown in Table 4-6, fine-tuning CNN layers, fine-tuning BN layers and fine-tuning both of them can get better results than those without fine-tuning the weights in these two kinds of layers. The smaller the data set, the better performance of fine-tuning can be improved. This indicates that fine-tuning CNN layers or BN layers are effective for ferrograph image classification, especially for the small data set. Moreover, the results of fine-tuning BN layers are better than those of fine-tuning CNN layers or fine-tune both layers in most cases. Besides, the results of fine-tune both layers are better than those of fine-tune CNN layers. This shows that the BN layers need to be fine-tuned for ferrograph image classification tasks. A surprising result is that fine-tune BN layers achieve better results than those of fine-tune both CNN and BN layers in most cases. Note that different learning rates are used in these two fine-tuning methods. Therefore, this surprising result may be caused by different learning rates. In order to validate this conclusion, the following experiments are implemented. We set different learning rate in different layers, that is, we set learning rate 0.01, 0.001and 0.0001 in BN layers, FC layers and CNN layers respectively. The results are shown in Table 7.

Table 7. Different learning rate in different kinds of layers VS the same learning rate in different kinds of layers

| Model | Data set | Learning rate 0.0001 for CNN, BN & FC layers | Learning rate 0.01 for BN & FC layers | Learning rate 0.01, 0.001, 0.0001 for BN, FC & CNN layers |
|---|---|---|---|---|
| DenseNet121 | FI140 | 66.4018 | 66.9694 | 70.6016 |
| DenseNet201 | FI140 | 67.7639 | 69.2395 | 71.3961 |
| ResNet18 | FI140 | 64.5857 | 66.2883 | 67.6504 |
| ResNet50 | FI140 | 65.0397 | 66.8558 | 72.3042 |
| ResNet152 | FI140 | 66.7423 | 65.1532 | 72.1907 |
| DenseNet121 | FI280 | 68.5585 | 83.0874 | 84.6765 |
| DenseNet201 | FI280 | 71.1691 | 82.7469 | 80.3632 |
| ResNet18 | FI280 | 70.2611 | 80.0227 | 78.6606 |
| ResNet50 | FI280 | 74.3473 | 81.0443 | 83.4279 |
| ResNet152 | FI280 | 72.1907 | 79.2281 | 80.2497 |
| DenseNet121 | FI909 | 90.2384 | 93.7571 | 94.0976 |
| DenseNet201 | FI909 | 91.941 | 93.4166 | 94.4381 |
| ResNet18 | FI909 | 87.9682 | 90.3519 | 92.5085 |
| ResNet50 | FI909 | 89.7843 | 93.076 | 93.6436 |
| ResNet152 | FI909 | 91.0329 | 92.2815 | 93.4166 |

It can be found in Table 7 that fine-tune different kinds of layers with different learning rate achieve a better result in most cases. However, its' improvements are small compared with fine-tuning BN layers. Therefore, fine-tune BN layers is an efficient approach. Since we need adopt a large learning rate to update BN layer, it means the parameters of the BN layers in the pre-train models on ImageNet are may vary greatly with those in the well convergent models on ferrograph image data set. Hence, the parameters of the BN layers may relate to different kinds of datasets, that is, it may reflect the semantic differences between different types of data sets. We will validate this conclusion in Section 3.4.

### 3.3 Fine-tuning different BN layers

The experiment results in Section 3.1 and Section 3.2 show that fine-tuning layers achieves good results. Reference [10] shows that different CNN layers obtain different features. The shallow layers learn general features while the deep layers learn semantic features. Therefore, it is a feasible and effective way to only fine-tine the deep layers of a model for small data sets. In order to explore whether the fine-tuning BN layers has the same effect, we implement experiments of updating different BN layers. The results are shown in Table 8.

As shown in table 8, the more BN layers updated, the better results are obtained. Besides,

updating the first BN layer also achieves greater performance improvement.

Table 8. Fine-tune different BN layers

| Models | Data set | Layer 4 | Layer 3 & 4 | Layer 2, 3 & 4 | All BN layers |
|---|---|---|---|---|---|
| Densenet121 | 140 | 60.2724 | 60.7264 | 62.5426 | 66.9694 |
| Resnet50 | 140 | 62.4291 | 66.8558 | 67.3099 | 66.8558 |
| Densenet121 | 280 | 66.4018 | 67.8774 | 70.3746 | 83.0874 |
| Resnet50 | 280 | 71.5096 | 75.3689 | 77.7526 | 81.0443 |
| Densenet121 | 909 | 78.6606 | 84.2225 | 86.4926 | 93.7571 |
| Resnet50 | 909 | 85.6981 | 89.5573 | 92.0545 | 93.076 |

## 3.4 Weights divergence analysis

In section 3.1-3.3, we carry out a large number of experiments. The results show that fine-tune the CNN layers or the BN layers obtain better results than those learning from scratch. In addition, fine-tuning BN layers are superior to fine-tuning CNN layers. Moreover, fine-tuning more BN layers obtain better results. In order to explore the possible reasons for these results, we further apply the formula (1) to analyze the weights difference of different well convergent models.

We will analyze the difference between the parameters in CNN layers with fine-tuning and pre-trained models provided by ImageNet. The details are as follows: we first fine-tune the CNN layers until the model is convergent. Then, the CNN weights of the convergent model will be compared to the weight of the original pre-trained model. In this study, the ResNet50 model and DenseNet121 model are used for experimental analysis. The experimental results are shown in Figure 1 and Figure 2.

As shown in Figure 1-2, the CNN weights divergence of the shallow layers is small, while CNN weights divergence of the deep layers is large. This means that the shallow layers learn the general features while the weights in deep layers are closely related to specific tasks. Thus, the weight divergence of the shallow layers is small. Besides, the CNN weights divergence analysis on different ferrograph image data sets shows that the weights difference increases with the increase of data quantity difference. In addition, the larger number of ferrograph image, the greater the difference between the fine-tuned weights and weights in pre-trained models. Combined with the experimental results in Section 3.3, it is shown that when the number of the ferrograph image increase, the weights in the model will make a large adjustment to the weights in the pre-trained model, and thus the fine-tuned model can learn the distribution of data better and achieve lower test error rate.

Similarly, we analyze the parameters (both the weights and bias) divergence of BN layers. The weights divergence results are shown in Figure 3-4 and the bias divergence results are shown in Figure 5-6. It can be seen from Figure 3-6 that the parameters of the BN layers after fine-tuned are greatly different from that of the original parameters of the pre-trained BN layers. While the parameters of the BN layers after fine-tuned on different ferrograph image data set is a little different. This indicates the parameters of the BN layer are related to the semantic information of the data set. In addition, the parameter divergence of the BN layers is much greater than that of the CNN layers. This phenomenon supports the argument in Section 3.3 that the BN layer should adopt a big learning rate, while the convolution layer should adopt a small learning rate to fine-tune the model. Moreover, parameter divergence in different BN layers seems to be the same, which verifies the experimental results in Section 3.3 that fine-tune all BN layers will get better results. What's more, the bias difference of the last BN layer is relatively large in most cases, which indicates that the bias of the last BN layer has strong links with the specific task.

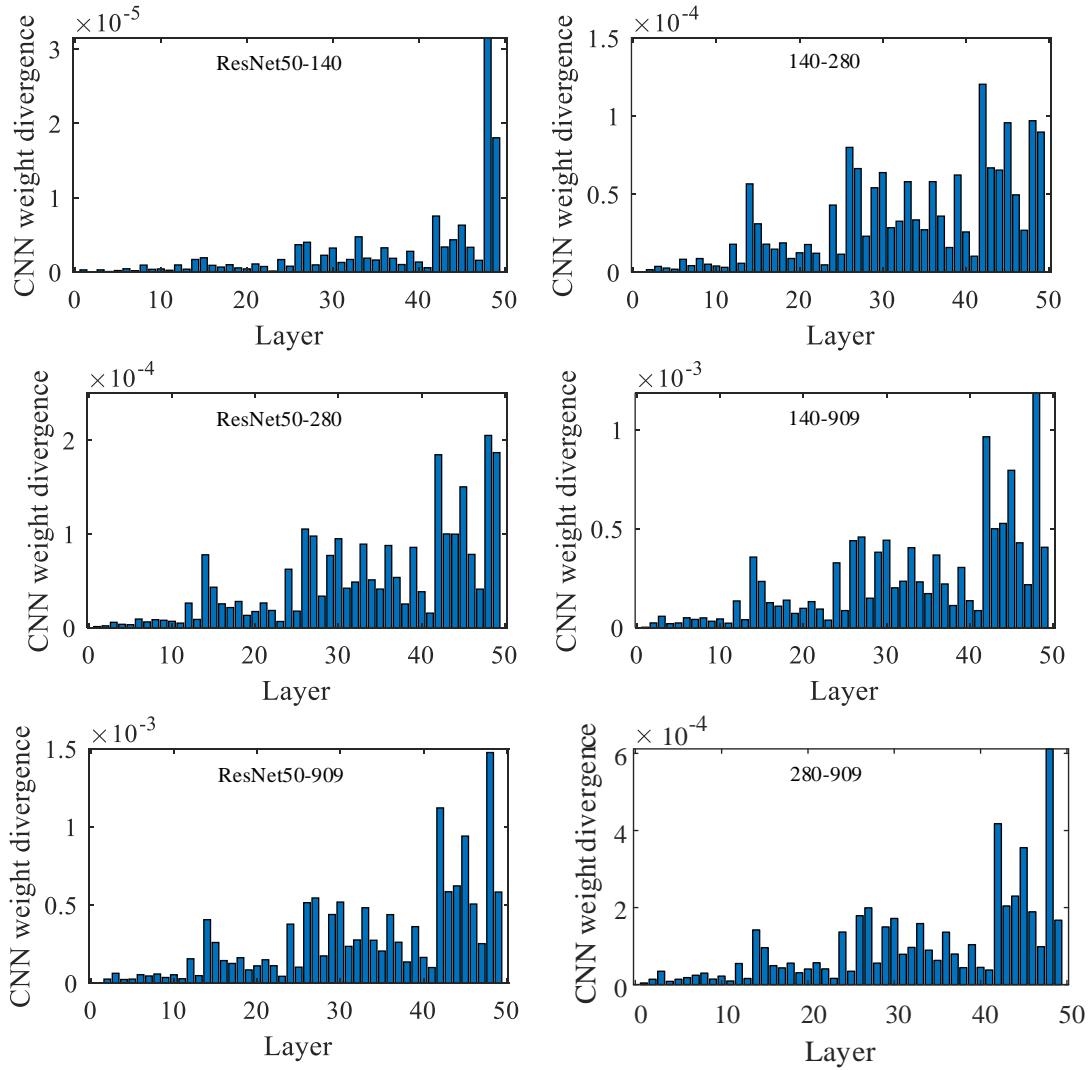

Figure 1. CNN weights divergence in different ResNet50 trained with different data sets. The figure with the title 'ResNet50-140' indicates this figure shows the CNN weights divergence between the fine-tuned model on FI140 and the pre-trained model, and the figures with 'ResNet50-280' and 'ResNet50-909' are extrapolated in this way. The Figure with the title '140-280' means this figure shows the CNN weights divergence between two fine-tuned models on FI140 and FI280, and the figures with '140-909' and '280-909' are extrapolated in this way.

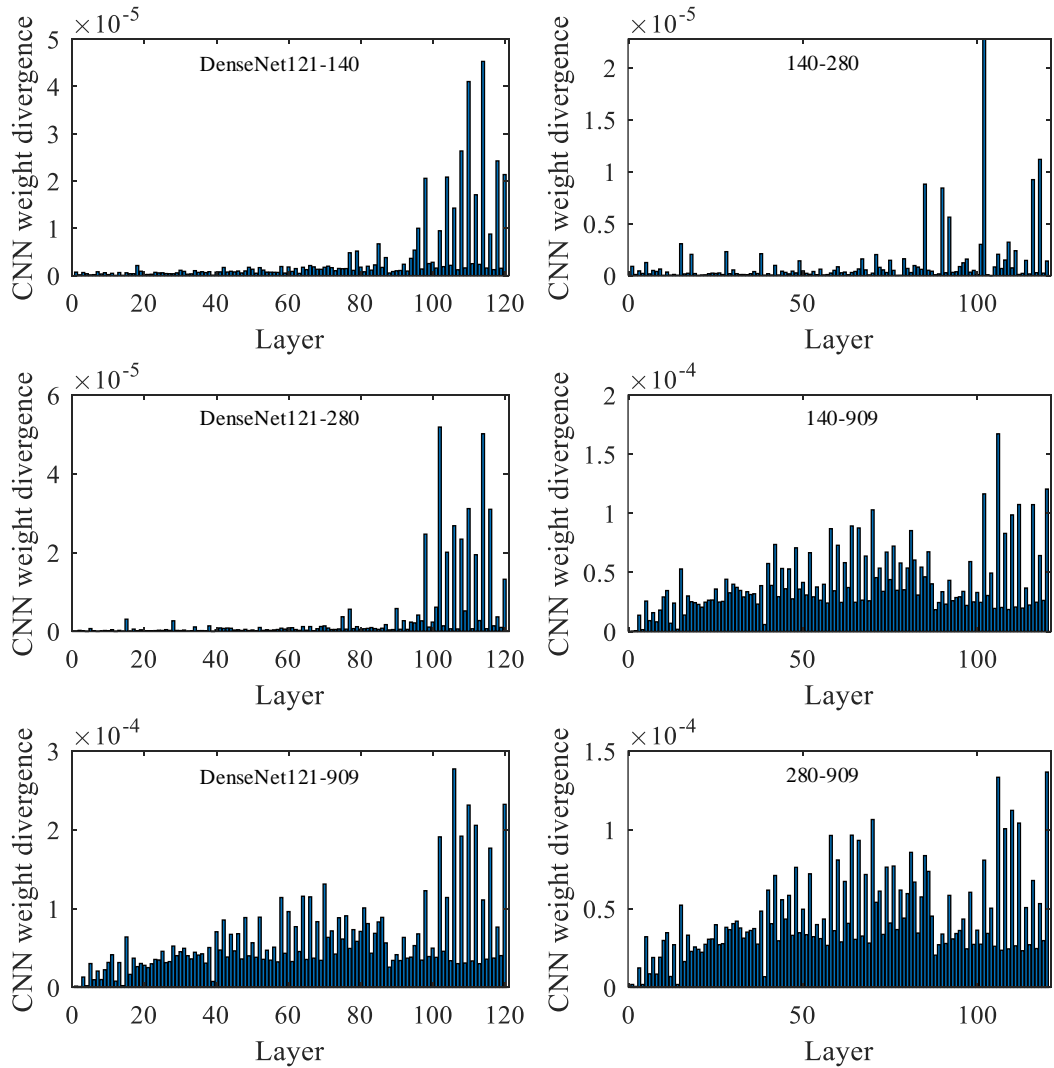

Figure 2. CNN weights divergence in different DenseNet121 trained with different data sets. The figure with title 'DenseNet121-140' indicates this figure shows the CNN weights divergence between the fine-tuned model on FI140 and the pre-trained model, and the figures with 'DenseNet121-280' and 'DenseNet121-909' are extrapolated in this way. The Figure with the title '140-280' means this figure shows the CNN weights divergence between two fine-tuned models on FI140 and FI280, and the figures with '140-909' and '280-909' are extrapolated in this way.

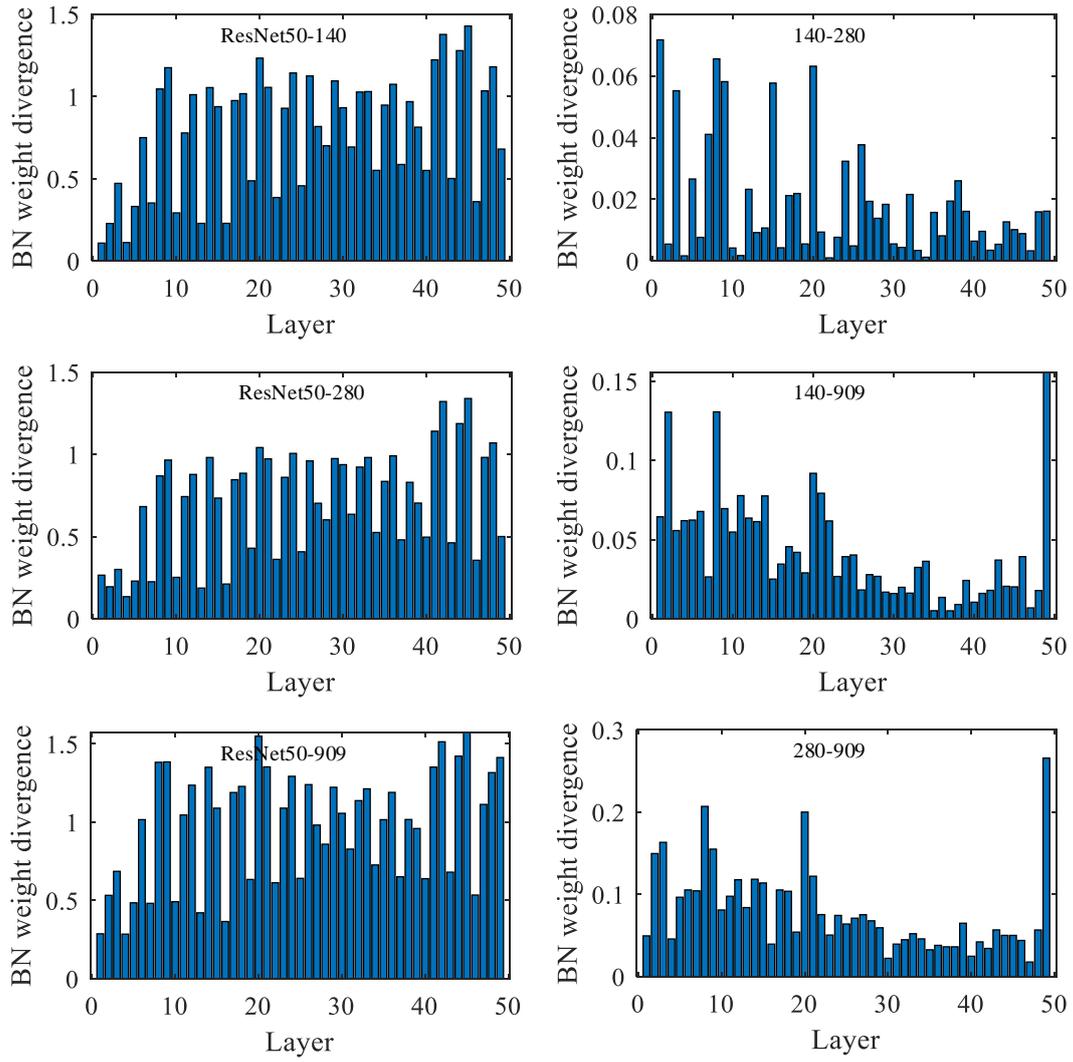

Figure 3. BN weights divergence of ResNet50.

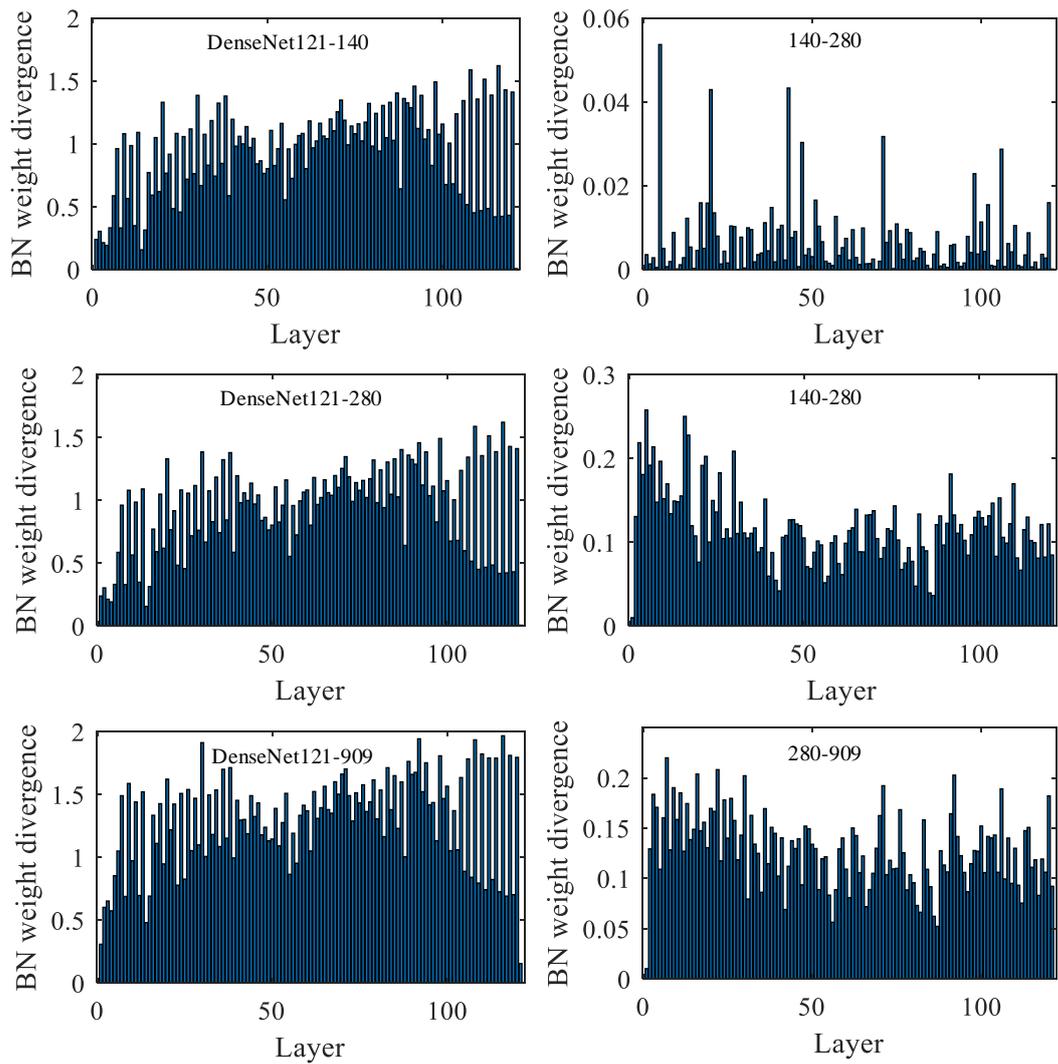

Figure 4. BN weights divergence of DenseNet121.

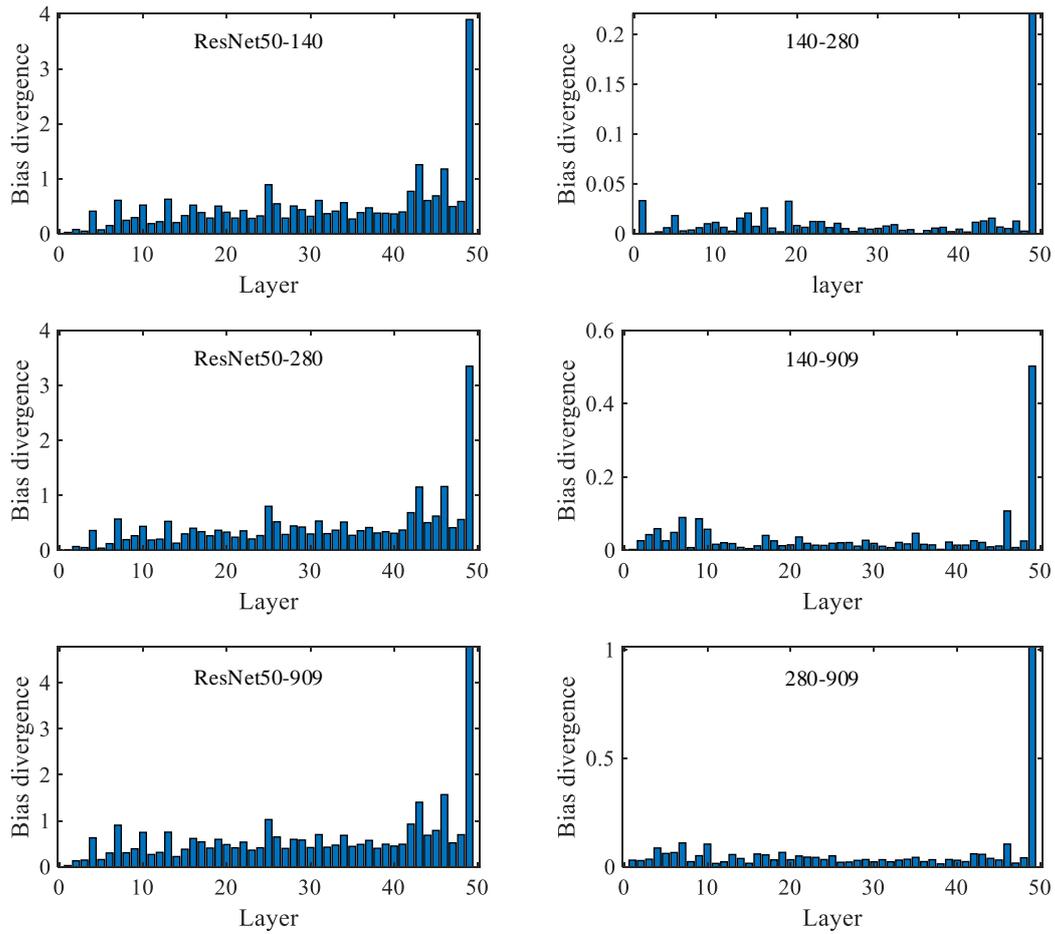

Figure 5. BN bias divergence of ResNet50.

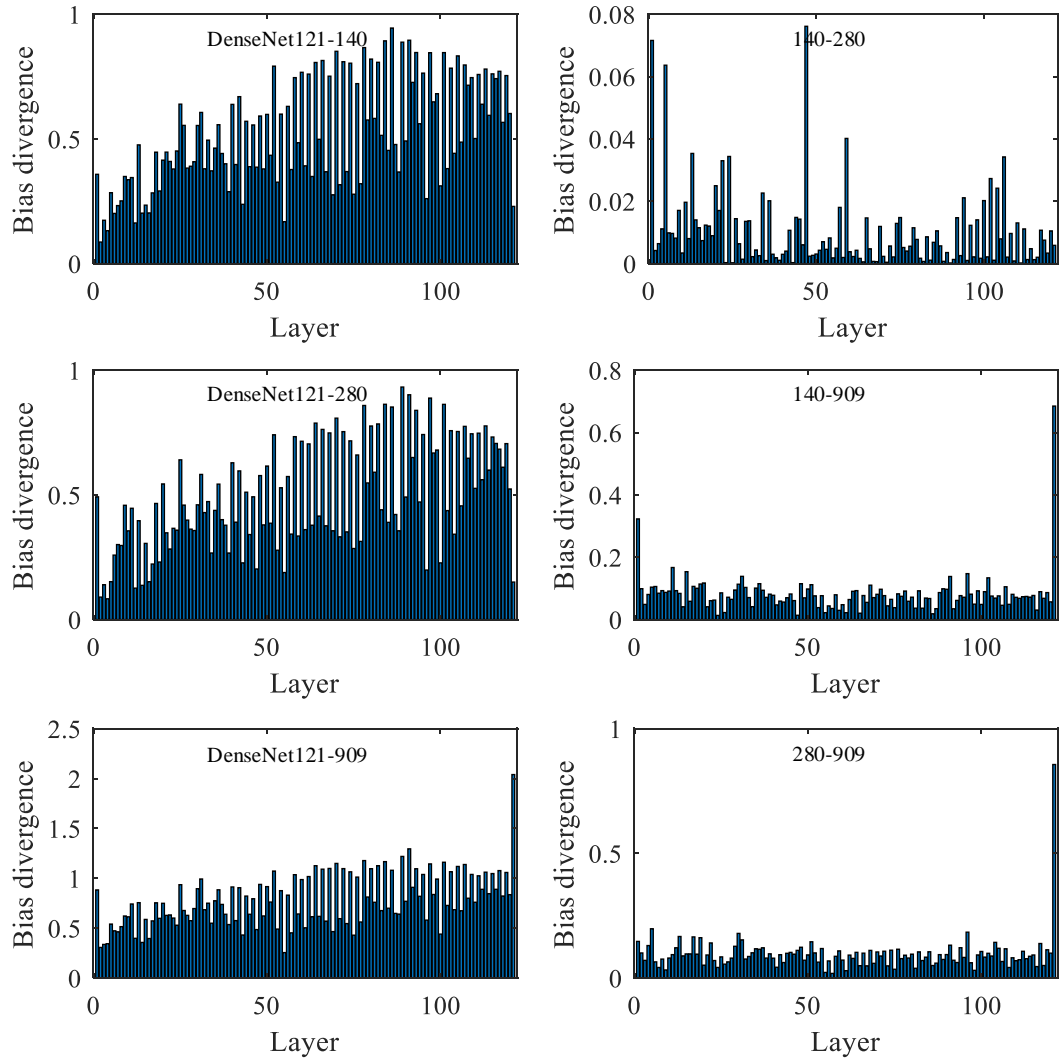

Figure 6. BN bias divergence of DenseNet121.

## Conclusion

We have done a large number of experiments on fine-tuning, and the parameter difference between the fine-tuned model and pre-train model are analyzed. We can obtain the following conclusions in this study.

(1) We can fine-tune the ImageNet pre-training model to achieve the classification of ferrograph images. The less amount of ferrograph images, the greater improvement of fine-tuning.

(2) Parameters in the BN layers have strong links with the semantic information of different data set. Thus, the BN layers need to be updated if the target domain and the source domain have different semantics. Besides, all BN layers should be updated.

(3) Most experimental results show that fine-tuning BN layers is better than fine-tuning CNN layers. In addition, setting different learning rates at different kinds of layers can further improve the performance.

(4) The results of weight difference analysis show that the weights of the shallow CNN layers are smaller than those of the deep layers. However, the weight of the BN layers does not meet this rule, and the weights of all the BN layers have great differences.

# Reference

[1] Krizhevsky A, Sutskever I, Hinton G. ImageNet classification with deep convolutional neural networks. Communications of the ACM, 2017. 60(6): p. 84-90.

[2] Simonyan K and Zisserman A. Very Deep Convolutional Networks for Large-Scale Image Recognition. arXiv preprint arXiv:1409.1556, 2014.

[3] Szegedy C, Liu W, Jia Y. et al. Going deeper with convolutions. In Proceedings of the IEEE Conference on Computer Vision and Pattern Recognition, pages 1–9, 2015.

[4] Ioffe, S. and C. Szegedy. Batch Normalization: Accelerating Deep Network Training by Reducing Internal Covariate Shift. arXiv preprint arXiv:1502.03167, 2015.

[5] Szegedy C, Vanhoucke V, Ioffe S. et al. Rethinking the Inception Architecture for Computer Vision. 2016 IEEE Conference on Computer Vision and Pattern Recognition, 2016:2818-2826.

[6] Szegedy C, Ioffe S, Vanhoucke, V. et al. Inception-v4, Inception-ResNet and the Impact of Residual Connections on Learning. AAAI Conference on Artificial Intelligence, 2016.

[7] He K, Zhang X, Ren S. et al. Deep Residual Learning for Image Recognition. IEEE Conference on Computer Vision & Pattern Recognition. 2016.

[8] Huang G, Liu Z and Weinberger K. Densely connected convolutional neural networks. arXiv preprint arXiv:1608.06993, 2017.

[9] Wang Y, Yao Q, Kwok J. et al. Generalizing from a Few Examples: A Survey on Few-shot Learning. ACM computing surveys, 2020. 53(3): p. 1-34.

[10] Yosinski J, Clune J, Bengio Y. et al. How transferable are features in deep neural networks? International Conference on Neural Information Processing Systems. MIT Press, 2014.

[11] Pan X, Luo P, Shi J. et al. Two at Once: Enhancing Learning and Generalization Capacities via IBN-Net. arXiv preprint arXiv:1807.09441v3, 2020.

[12] Roylance B. Ferrography—then and now. Tribology International, 2005. 38(10): p. 857-862.

[13] Wu T, Mao J, Wang J. et al. A New On-Line Visual Ferrograph. Tribology Transactions, 2009. 52(5): p. 623-631.

[14] Peng Y, Cai J, Wu T. et al. A hybrid convolutional neural network for intelligent wear particle classification. Tribology International, 2019. 138: p. 166-173.

[15] Peng P. and Wang J. Wear particle classification considering particle overlapping. Wear, 2019. 422-423: p. 119-127.

[16] Pan S, Yang Q. A Survey on Transfer Learning. Knowledge and Data Engineering. 2010. 22: p1345 - 1359.

[17] Li Y, Wang N, Shi J. et al. Adaptive Batch Normalization for practical domain adaptation. Pattern Recognition, 2018. 80: p109-117.